%% file: root.tex
\documentclass[12pt, final]{l4dc2020} 

\title[Learning Constrained Dynamics with Gauss' Principle adhering Gaussian Processes]{Learning Constrained Dynamics\\  with Gauss' Principle adhering Gaussian Processes}
\usepackage{times}
\usepackage{graphics} % for pdf, bitmapped graphics files
\usepackage{amsmath} % assumes amsmath package installed
\usepackage{amssymb}  % assumes amsmath package installed
\usepackage{comment}
\usepackage[capitalise]{cleveref}
\usepackage{bm}
\usepackage{xspace}
\usepackage{xcolor}
\usepackage{float}
\usepackage{multirow}
\usepackage{caption}
\usepackage{subcaption}
\usepackage{wrapfig}
\usepackage{appendix}
\usepackage{siunitx}
\usepackage{booktabs}
\usepackage{pdfpages}

\newtheorem{exmp}{Ex.} %[section]

\newcommand{\transp}{\mathrm{\sf T}}  % Seb: changed notation for transpose
\newcommand{\E}{\mathbb{E}}

\newcommand{\eg}{\emph{e.g.},\xspace}

\newcommand\colorrulemix[1]{\textcolor{#1!40!black}}
\newcommand{\change}[1]{\setlength{\fboxsep}{0pt}\colorbox{white!100!white}{#1}}  %my notes
\author{%
 \Name{A. Ren\'e Geist}$\: ^{\dag}$ \Email{geist@is.mpg.de}\\
 \Name{Sebastian Trimpe}$\: ^{\dag}$ \Email{trimpe@is.mpg.de}\\
 \addr $^{\dag}$\: Intelligent Control Systems Group, Max Planck Institute for Intelligent Systems
}

\begin{document}

\maketitle

\begin{abstract}%
The identification of the constrained dynamics of mechanical systems is often challenging. 
Learning methods promise to ease an analytical analysis, but require considerable amounts of data for training. We propose to combine insights from analytical mechanics with Gaussian process regression to improve the model's data efficiency and constraint integrity. The result is a Gaussian process model that incorporates a priori constraint knowledge such that its predictions adhere to Gauss' principle of least constraint. In return, predictions of the system's acceleration naturally respect potentially non-ideal (non-)holonomic equality constraints. As corollary results, our model enables to infer the acceleration of the unconstrained system from data of the constrained system and enables knowledge transfer between differing constraint configurations.
\end{abstract}

\begin{keywords}%
  Constrained Lagrangian systems, Gauss' Principle, Gaussian Processes, Nonlinear system identification, Structured learning, Transfer learning
\end{keywords}

%===============================================================================
\section{Introduction}
The acquisition of accurate models of dynamical systems is essential for a multitude of engineering applications. If the function generating the data is unknown or too complex to be modelled from first principles, non-parametric learning models aim at inferring a function solely from the data. Gaussian Processes (GPs) are non-parametric and have been commonly used to learn dynamics. They have demonstrated their versatility in approximating continuous nonlinear functions on a plethora of real-world problems \citep{williams2006gaussian}, including modeling of dynamical systems \citep{nguyen2011model,kocijan2005dynamic,frigola2013bayesian, mattos2016latent, doerr2017optimizing, eleftheriadis2017identification, doerr2018probabilistic}. In this context, GPs are often preferred over alternative methods, since they provide a measure of the uncertainty about function estimates in the form of the posterior variance. Additionally, they allow for the incorporation of various model assumptions through the covariance function (kernel). However, data on real-world systems is often scarce and contains only partial information, which is why training purely data-driven models
%may still require large amounts of data to obtain a
to sufficient prediction accuracy is challenging. Further, predictions made by standard GP models may %extrapolate poorly to unobserved parts of the function.
violate critical system constraints compromising the predictions' integrity.

A promising approach to improve the data efficiency and constraint integrity of a GP model involves the incorporation of a priori available \textit{structural knowledge} in the design of the covariance function. Turning to mechanical systems, the identification of fundamental structural relationships has been studied intensively by scholars for the last two centuries under the subject of analytical mechanics. In this work, we leverage the structure inherent in holonomic and non-holonomic constraint equations, which also can be non-ideal. For example, a pendulum's rod or a rolling wheel enforce a holonomic or non-holonomic constraint, respectively. A constraint is referred to as being \emph{non-ideal} if it induces forces onto the system that produce \change{virtual} work (\eg damping and friction), and it is called \emph{ideal} if no \change{virtual} work is produced (\eg pendulum rod). The constrained dynamics of such systems are described by the Udwadia-Kalaba equation (UKE) \citep{udwadia1992new,udwadia2007analytical}. The UKE is a direct result of \emph{Gauss' principle} of least constraint \citep{gauss1829neues}, which states that a system's constrained acceleration can be cast as the solution of a least-squares problem. %on the accelerations. 

In this work, we propose a GP model that leverages mechanical constraints as prior knowledge for learning dynamics of mechanical systems.  Specifically, a GP is transformed by constraint equations to satisfy Gauss’ Principle. The resulting \emph{Gauss' Principle adhering Gaussian Process} (GP\textsuperscript{2}) performs inference in a physically substantiated sub-space of the acceleration space. In return, the data-efficiency and physical integrity are improved compared to the untransformed GP. %as we show in numerical examples.

%In this work, we leverage that a GP transformed by mechanical constraint equations to fulfill Gauss' Principle performs inference in a physically substantiated sub-space of the acceleration space. In return, the data-efficiency and physical integrity of such a Gauss' Principle adhering Gaussian Process (GP\textsuperscript{2}) is improved compared to an untransformed GP, as we show in numerical examples.

\section{Problem Formulation} \label{sec:problemformulation}
The acceleration of constrained mechanical rigid-body systems is described as
\begin{equation}
    \ddot{q}=h(q,\dot{q}, t) = M^{-1}(q, t) F(q,\dot{q}, t), \quad h : \mathbb{R}^{D} \to \mathbb{R}^{n}
    \label{eq:dynamics_h}
\end{equation}
%$\ddot{q}=h(q,\dot{q}): \mathbb{R}^{D} \mapsto \mathbb{R}^{n}$,
with state dimension $n$, $D=2n+1$, symmetric positive definite matrix $M(q, t)$, and vector $F(q,\dot{q}, t)$. 
%where the dimensions ($n$ and $D=2n+n_u$) and positive definite-inertia matrix $M(q)$ are assumed to be known, but the forces $F_a$ and $F_{\tau}$ are \emph{unknown}. 
The variables $q$, $\dot{q}$, $\ddot{q}$ correspond to the system's positions, velocities,  and accelerations, respectively. Although they depend on time $t$, we generally omit the time dependence for these and other variables if this is clear from the context. Inhere, the system is solely subject to constraint forces arising from sufficiently smooth \emph{(non-)holonomic constraints}, $c_i(q, \dot{q}, t,\theta_p)=0$, with $i=1,2,...,m$, whose (second) time-derivatives are linear in $\ddot{q}$, yielding
\begin{equation} \label{eq:constraint}
    A(q,\dot{q},t)\ddot{q} = b(q,\dot{q},t),
\end{equation}
with functions $A : \mathbb{R}^{D} \to \mathbb{R}^{m \times n}$, $b : \mathbb{R}^{D} \to \mathbb{R}^{m}$, and $m<n$.
We refer to \eqref{eq:constraint} as the \emph{constraining equation}.
Most applications of analytical mechanics deal with such constraints \citep[p.~80]{udwadia2007analytical}. These constraints can be \emph{non-ideal}, but naturally must be consistent.

Further, it is assumed that the parametric functions $\{A,b,M\}$ are known, \eg from a preceding mechanical analysis of the ``unconstrained'' rigid-body dynamics (see \cref{sec:constrainted-modeling-mech-sys}), while the system's parameters $\theta_p = [p_1,...,p_r]$ are potentially unknown. For the above-defined system, data $\mathcal{D}=\{x_k,y_k\}^{N}_{k=1}$ is available, consisting of input points $x_k = [q_k, \dot{q}_k, t_k]^{\transp}$ and observations $y_k = h(x_k) + \epsilon_k$ where $y_k \in \mathbb{R}^{n}$ with $\epsilon_k$ denoting zero-mean Gaussian noise.  In some cases, $x_k$ includes a vector of control forces $u_k\in\mathbb{R}^{n_{u}}$ explicitly. While more general settings such as hidden states can be addressed in GP dynamics learning \citep{doerr2018probabilistic}, we here focus on the standard GP regression setting with noiseless inputs and noisy targets.

The main objective of this work is to learn $h$ via a GP $\hat{h}\sim\mathcal{GP}(\mu_{\hat{h}},K_{\hat{h}})$ that incorporates the structural knowledge $\{A,b,M\}$ in the GP's mean $\mu_{\hat{h}}(x|\theta_p)$ and covariance function $K_{\hat{h}}(x,x'|\theta_p)$ such that its posterior mean approximates \eqref{eq:dynamics_h} while satisfying \eqref{eq:constraint} exactly.

%===============================================================================
\section{Related Work}
We categorize related literature into the following aspects: \emph{(i)} leveraging functional relations arising from analytical mechanics to structure learning; \emph{(ii)} constructing GP kernels using semi-parametric models; and \emph{(iii)} using linear transformations to provide structure to GP models.

The incorporation of functional relationships arising from the field of analytical mechanics starts gaining attention.
For example, \citep{ledezma2018fop} use the recursive nature of the Newton-Euler equations of open-chain rigid-body configurations to derive a parametric regression model.
Recently, \citep{lutter2018deep} detailed the direct incorporation of the Lagrange equations describing the inverse dynamics of \change{holonomic ideal constrained} systems into neural networks. Further, \citep{greydanus2019hamiltonian} combined Hamiltonian mechanics with neural networks to predict the forward dynamics of conservative mechanical systems. Our work is linked to \citep{cheng2015learn}, in which the operators underlying the Lagrange equations are used to derive kernels which capture inverse dynamics as the Lagrangian’s projection. In comparison, our model builds on projection operations underlying mechanical constraint equations to enable structured learning of forward dynamics on possibly non-holonomic and non-ideally constrained systems.

If a parametric function is cast as a (Bayesian) linear regression problem, 
a degenerate GP covariance function can be derived to model the function non-parametrically.
This relationship is used in \citep{nguyen2010using} to derive a structured GP kernel for learning the inverse dynamics of open-chain robot arms.
In our work, we do not assume that the dynamics are linear in the system parameters, but instead, leverage that \eqref{eq:constraint} is linear with respect to $\ddot{q}$.

Because GPs are closed under linear operations, and linear operators commonly occur in physical equations, such operators play a particularly important role in structured learning with GPs. Linear operations commonly used in literature are, \eg  differentiation \citep{solak2003derivative}, integration, and wrapping of the GP inputs into a nonlinear function \citep{calandra2016manifold}. Furthermore, one particular application of GP regression revolves around efficiently learning the solution of differential equations by transforming GPs through convolution operators \citep{alvarez2009latent, sarkka2011linear}. \citep{jidling2017linearly, lange2018algorithmic} discuss how to construct a GP kernel such that its realizations $f(x)$ fulfill a constrained equation of the form $A_x f(x)=0$, where $A_x$ is a linear operator. % (in particular elements of a polynomial ring or Weyl algebra).  
While our work also transforms a GP such that its predictions lie in a linear operator's nullspace, we emphasize that in classical mechanics such operators enable the construction of a GP whose predictions satisfy nonlinear constraints in a physically meaningful manner.
In comparison, the work in \citep{agrell2019gaussian} considers modeling a GP to fulfill an inequality constraint of the form $A_xf(x) \leq b(x)$ by conditioning its posterior on carefully selected virtual observations of $b(x)$.  In contrast to \citep{jidling2017linearly, lange2018algorithmic} and this work, \citep{agrell2019gaussian} ensures constraint satisfaction with a certain probability at selected sample points, while we ensure constraint satisfaction over the whole input range.

None of the above works addresses the problem of constrained modeling of rigid-body  mechanical systems through a tailored GP satisfying an affine equality constraint as stated in \cref{sec:problemformulation}.

%===============================================================================
\section{Constrained Dynamics with Gauss' Principle}
\label{sec:constrainted-modeling-mech-sys}
In this section, we present a description of constrained mechanical systems that gives an explicit form for the constraint forces acting on a system as a function of the state $x$ and constraining equation (2) as detailed in \citep{udwadia2002foundations} and Sec.\,S3 and S4 of the supplementary material.  This allows us in the following section to construct a GP that respects the constraints underlying (2).
%

\begin{comment}
\begin{wrapfigure}{r}{50mm} 
  \vspace{-10pt}
  \begin{center}
    \resizebox{50mm}{!}{\input{fig/tangent_on_surface_small.pdf_tex}}
  \end{center}
  \vspace{-15pt}
  \caption{A mass particle sliding along a surface according to Gauss' principle. \label{fig:surface_particle}}
\end{wrapfigure}
%
\end{comment}
\begin{comment}
\begin{figure}
\vspace{-1cm}
\centering
    \begin{minipage}[b]{0.4\textwidth}
    \resizebox{50mm}{!}{\input{fig/tangent_on_surface_small.pdf_tex}}
  \caption{A mass particle sliding along a surface according to Gauss' principle. \label{fig:surface_particle}}
    \end{minipage}%
      \hspace{1cm}
    \begin{minipage}{.4\textwidth}
  \vspace{-172pt}
    \resizebox{50mm}{!}{\input{fig/unicycle2.pdf_tex}}
  \resizebox{50mm}{!}{\input{fig/duffing.pdf_tex}}
%\includegraphics[width=0.5\linewidth]{fig/duffing.pdf_tex}
  \captionof{figure}{Unicycle (top) and controlled Duffing oscillator (bottom).}
  \label{fig:benchmark_systems}
    \end{minipage}%
    \vspace{-0.5cm}
\end{figure} 
\end{comment}

\begin{figure}
    \begin{minipage}[b]{0.37\textwidth}
    \centering
    \resizebox{50mm}{!}{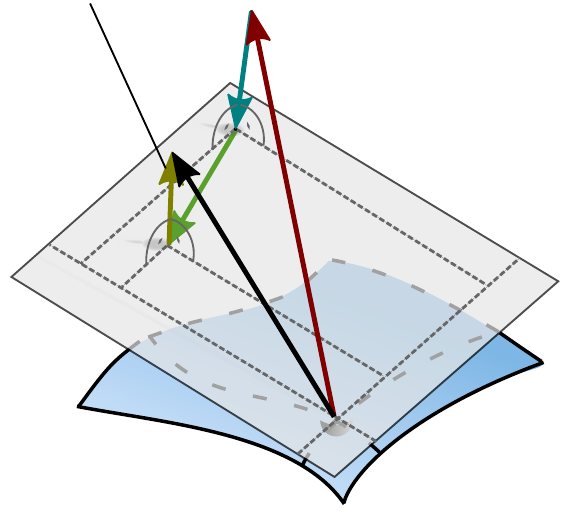}
        \captionof{figure}{Mass particle sliding on a surface subject to Gauss' principle.}
        \label{fig:surface_particle}
    \end{minipage}
        \hspace{0.4cm}
    \begin{minipage}[b]{0.25\textwidth}
    \centering
        \resizebox{28mm}{!}{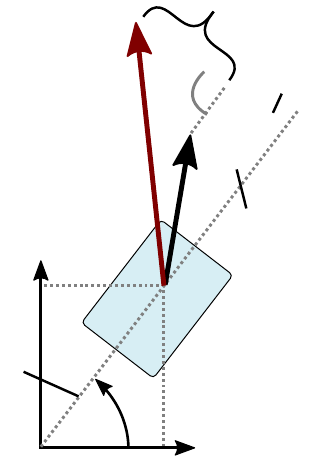}  
        \captionof{figure}{Unicycle subject to Gauss' Principle.}
        \label{fig:unicycle}
    \end{minipage}%
    \hspace{0.4cm}
  \begin{minipage}[b]{0.28\textwidth}
  \centering
        \resizebox{55mm}{!}{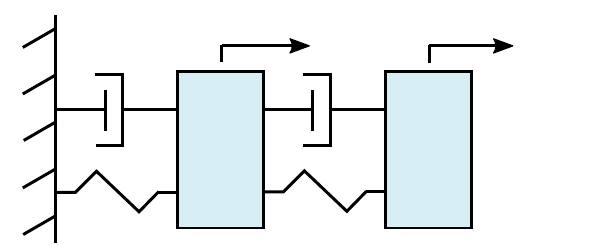}
        \captionof{figure}{Duffing oscillator. \\ \hspace{0.1cm}}
        \label{fig:duffing}
    \end{minipage}
    \vspace{-0.2cm}
\end{figure} 

We consider mechanical systems where a force $F_a$ acting on rigid bodies results in the  \emph{unconstrained acceleration} 
\begin{equation}
    a = M(q,t)^{-1}F_a(x). 
\end{equation}
\begin{comment}
The term ``unconstrained'' refers either to $a(x)$ being solely described by \change{independent} coordinates, which in return does not require the imposition of external constraint equations to describe the system's motion, \emph{or}, $a(x)$ is describing constrained Lagrangian dynamics onto which an additional constraining equation \eqref{eq:constraint} has not yet been applied.
\end{comment}
The term ``unconstrained'' refers to $a(x)$ describing potentially constrained Lagrangian dynamics onto which an (additional) constraining equation \eqref{eq:constraint} has not yet been applied.
If an (additional) constraint acts on the system,
its movement changes to the \textit{constrained acceleration} $\ddot{q}=a+\tau$. We refer to the term $\tau = \tau_{\text{ideal}}+z$ as the \emph{constraining acceleration}, where $z(x)$ denotes the non-ideal part of $\tau(x)$ (\eg damping and friction). \cite{udwadia1997equations} refers to a constraining acceleration as non-ideal if it lies in the nullspace of $A(x)$, writing $\mathcal{N}(A)$, and as ideal if it lies in the range space of $A(x)$.  \cite{gauss1829neues} observed that the \emph{ideal constraining acceleration}  $\tau_{\text{ideal}}(x)$ minimizes the functional $G(x) = \tau_{\text{ideal}}^{\transp}M\tau_{\text{ideal}}$. This fundamental principle underlying the constrained motion of rigid-body systems is referred to as \emph{Gauss' principle of least constraint}. The minimizing solution of this least-squares problem is given by the UKE -- omitting the dependencies on $x$ for clarity -- as
\begin{align}
    \ddot{q} = \underbrace{M^{-1}A^{\transp}(AM^{-1}A^{\transp})^{+}}_\text{\normalsize $L(x,\theta_p)$}b + \underbrace{(I-M^{-1}A^{\transp}(AM^{-1}A^{\transp})^{+}A)}_\text{\normalsize $T(x,\theta_p)$}\underbrace{(a+z)}_\text{\normalsize$\bar{a}(x)$}, \label{eq:general-uke-long}
\end{align}
where  $(\cdot)^{+}$ denotes the Moore-Penrose pseudo (MP) inverse and $I$ is an identity matrix.
The UKE satisfies \eqref{eq:constraint} by construction and corresponds to \eqref{eq:dynamics_h}, for which we shall build a GP model. Next, we introduce three examples, which we use throughout for illustration. Further details can be found in Sec.\,S6 of the supplementary material and \citep[p.~120, p.~213]{udwadia2007analytical}.

\begin{exmp}[Particle on surface] \label{ex:example_surface}
Consider a particle as illustrated in \cref{fig:surface_particle} with the known mass $m$ sliding along a surface. While the dynamics of the particle are unknown, we want to leverage the surface geometry. The unconstrained acceleration of the particle amounts to $a=M^{-1}[u_1, u_2, u_3-mg]^{\transp}$ with $M^{-1}=\mathrm{diag}(1/m)$, $g=9.81\frac{\mathrm{m}}{\mathrm{s}^{2}}$, and control forces $u_i$. The mass slides on the surface $q_3 = p_1 q_1^2 + p_2 q_2^2 + p_3 q_1 + p_4\cos(p_5q_1)$, with the states and constraint parameters being denoted as \{$q_i,\dot{q}_i$\} and $\theta_p=[p_1,...,p_5]$. The second time-derivative of the constraint yields \eqref{eq:constraint} as
\begin{equation} \label{eq:particle_constraint}
\footnotesize
    \underbrace{\begin{bmatrix}2p_1q_1+p_3-p_4p_5\sin(p_5q_1), & 2p_2q_2, & -1\end{bmatrix}}_\text{$A(x,\theta_p)$} \ddot{q}=
    \underbrace{\begin{bmatrix}-2p_1\dot{q}_1^2-2p_2\dot{q}_2^2+p_4p_5^2\dot{q}_1^2\cos(p_5q_1)\end{bmatrix}}_\text{$b(x,\theta_p)$},
\end{equation}
In addition to $\tau_{\text{ideal}}$ resulting from \eqref{eq:particle_constraint}, a velocity quadratic damping force $F_z=Mz$ decelerates the mass non-ideally such that $z_i = - a_0(v^2/|v|)\dot{q}_i$, with the translatory velocity $v(x)$ and damping coefficient $a_0$. One obtains the system's constrained dynamics by inserting \eqref{eq:particle_constraint} as well as $a(x)$ and $z(x)$ into \eqref{eq:general-uke-long}. While $\theta_p$ can be readily measured and \{$A,b,M$\} are obtained from a brief mechanical analysis, modeling $F_a$ and $F_z$ pose a considerable challenge for a plethora of mechanical systems.
\end{exmp}

\begin{exmp}[\change{Unicycle}] \label{ex:example_unicycle}
The unicycle as depicted in \cref{fig:unicycle} commonly describes the motion of wheeled robots
\citep[p.\,478]{siciliano2010robotics}. Here, a non-holonomic constraint $\dot{q}_2=\dot{q}_1 \tan(q_3)$ only allows for instantaneous translation along the line C-G. 
%The distance between the points C and G amounts to $R$, and $I_c$ denotes the body's inertia in direction of $q_3$. 
Further, $u_1$ and $u_2$ denote control inputs in direction of C-G and around $q_3$, respectively. The system is decelerated in driving direction by $F_z$. $F_z$ is induced by velocity quadratic damping as detailed in \cref{ex:example_surface}.
\end{exmp}

\begin{exmp}[\change{Controlled Duffing's Oscillator}] \label{ex:example_duffing}
The \emph{Duffing's oscillator} as depicted in \cref{fig:duffing} models the behavior of two masses that are subject to cubic spring forces and liner damping. An external control force imposes the constraint $q_2=q_1+p_1\exp(-p_2 t)\sin(p_3 t)$. The uncontrolled system is defined as the unconstrained system such that $F_a$ origins from spring and damping forces.
\end{exmp}

\section{A Gaussian Process Model for Learning Constrained Dynamics}
\label{sec:GPmodel}
In this section, a constrained GP is derived from the UKE formulation of constrained mechanics in \cref{sec:constrainted-modeling-mech-sys}. To this end, we leverage that GPs are closed under linear transformations and that the operators underlying the UKE are projections. Thus, we obtain a transformed GP model for learning constrained dynamics whose mean and samples fulfill Gauss' principle.
%We also provide some insight into the transformations underlying the UKE, and why these lead to improved data efficiency of the resulting GP regression model.
%Afterwards, the transformations underlying the UKE are analyzed and the resulting properties of the GP regression model are discussed.

\subsection{Gaussian Process Regression}
\label{sec:GP-basics}
We consider learning of $h$ in \eqref{eq:dynamics_h} with GP regression.  Specifically, we approximate the true dynamics $h$ with a multi-output GP $\hat{h}$ \citep{alvarez2012kernels}, $\hat{h}(x) \sim \mathcal{GP}(\mu(x), K(x,x'))$
where $\mu : \mathbb{R}^{D} \to \mathbb{R}^{n}$ denotes the prior function mean, $\mu(x) = \E[h(x)]$, and $K(x,x') : \mathbb{R}^{D \times D} \to \mathbb{R}^{n \times n}$ the prior covariance, $K(x,x') \triangleq K_{x,x'} = \E[(h(x)-\mu(x)) (h(x')-\mu(x'))^\transp]$.
Given a GP model, predictions of $\hat{h}(x_{*})$ are made -- at a point $x_{*}$ using observations $y$ at inputs $X$  -- by computation of the covariance matrices $K(x_{*},X)$ and $K(X,X)$ and conditioning the GP via
\begin{align}
\mu_{x_{*}|X,y} &= \mu_{x_{*}}+ K_{x_{*},X}(K_{X,X}+\sigma_{y}^2I)^{-1}(y-\mu_{X}),  \label{eq:conditional_mean} \\ K_{x_{*}|X,y} &= K_{x_{*},x_{*}} - K_{x_{*},X}(K_{X,X}+\sigma_{y}^2I)^{-1}K_{x_{*},X}^{\transp}.
\end{align}

\subsection{Gauss Principle adhering Gaussian Processes}
\label{sec:UKE-GP-derivation}
The UKE \eqref{eq:general-uke-long} disentangles the acceleration $\ddot{q}(x)$ into a term that results from the unconstrained acceleration (being transformed), $T(x)a(x)$, one that is caused from the ideal part of the constraints \eqref{eq:constraint},  $L(x)b(x)$, and one caused by the non-ideal part of the constraints, $T(x)z(x)$. Hence, it provides the structure to build the sought GP model.  The second term is known from the structural knowledge $A(x,\theta_p)$, $b(x,\theta_p)$, and $M(x,\theta_p)$, while on the other terms, we will place a GP prior. More specifically, we consider two cases of prior structural knowledge: \emph{(i)} knowing the true $\theta_p=\theta_p^{*}$, or \emph{(ii)} knowing only the functional form of $A(x,\theta_p)$, $b(x,\theta_p)$, $M(x,\theta_p)$, but not the true parameters $\theta_p^{*}$. While we omit the dependencies on $\theta_p$ to ease the notation in the following, our work addresses both cases.  As for case \emph{(ii)}, $\theta_p$ will be treated as additional hyperparameters of the model.

We propose to model the unconstrained acceleration and non-ideal constraining acceleration jointly by placing a GP prior on $\bar{a}(x)$ such that $\hat{\bar{a}} \sim \mathcal{GP}(\mu_{\bar{a}}(x),K_{\bar{a}}(x,x'))$.
%The UKE \eqref{eq:general uke} represents a linear transformation of $\bar{a}$, $T(x)\bar{a}$, plus the offset term $L(x)b(x)$.  
As GPs are closed under linear transformations, inserting $\hat{\bar{a}}$ into \eqref{eq:general-uke-long} results in a GP modeling $\ddot{q}(x)$ as
\begin{equation} \label{eq:AMGP}
    \hat{h} \sim \mathcal{GP}\big(L(x)b(x)+T(x)\mu_{\bar{a}}(x),~T(x)K_{\bar{a}}(x,x')T(x')^{\transp}\big).
\end{equation}
To the above model, we refer to as a \textit{Gauss' Principle adhering Gaussian Process} (GP$^2$). %such that the constrained motion of a dynamical system is described by a parametric mean function and a transformed GP model. 
By construction, the GP\textsuperscript{2}'s predictions (mean and samples) satisfy \eqref{eq:constraint}. We further note the following favorable properties of the proposed model.

\paragraph{Flexible model.}
\change{If no prior} knowledge about the unconstrained dynamics is available $\mu_{\bar{a}}(x)$ can be set to be the null vector. \change{Alternatively,} if a prior model for the unconstrained dynamics is known, this knowledge can be incorporated in form of the mean function $\mu_{\bar{a}}(x)$ in \eqref{eq:AMGP}. This in return implies that $K_{\bar{a}}(x,x')$ solely models the non-ideal part of the constraint forces plus the residuals of the parametric model. For example, in \cref{ex:example_unicycle}, the functional structure of $a(x,\theta_p)$ is oftentimes known a priori. In this case, one can set $\mu_{\bar{a}}(x)=a(x,\theta_p)$. As we show in the experimental section, the  parameters describing $\mu_{\bar{a}}(x)$ can then be estimated alongside the parameters of $K_{\bar{a}}(x,x')$. %In return, the GP only approximates the change in the acceleration $z$ induced by the non-ideal friction and damping forces $F_z$.

\paragraph{Inferring the unconstrained acceleration alongside.}
As the constrained GP results from a linear transformation of $\bar{a}$(x), the joint distribution is obtained as
%
\begin{comment}
\begin{equation} \label{eq:joint distribution}
   \footnotesize \begin{bmatrix} 
   \bar{a} \\ 
   \tau \\ 
   \hat{h} \end{bmatrix} \sim \mathcal{GP} \begin{pmatrix} 
   \begin{bmatrix}  \mu_{\bar{a}}(x)\\ 
   L(x)b(x)+\bar{L}(x)\mu_{\bar{a}}(x)\\ 
   L(x)b(x)+T(x)\mu_{\bar{a}}(x) \end{bmatrix}, 
   \begin{bmatrix} 
   K_{\bar{a}}(x,x') & 
   K_{\bar{a}}(x,x')\bar{L}(x')^{\transp} & 
   K_{\bar{a}}(x) T(x')^{\transp} \\ 
   \bar{L}(x)K_{\bar{a}}(x,x') & 
   \bar{L}(x)K_{\bar{a}}(x,x')\bar{L}(x')^{\transp} & 
   \bar{L}(x)K_{\bar{a}}(x) T(x')^{\transp} \\
   T(x) K_{\bar{a}}(x) & 
   \bar{L}(x) K_{\bar{a}}(x)T(x')^{\transp} & T(x)K_{\bar{a}}(x,x')T(x')^{\transp}\end{bmatrix} \end{pmatrix}, \normalsize
\end{equation}
\end{comment}

\begin{equation} \label{eq:joint distribution2}
   \footnotesize \begin{bmatrix} \hat{\bar{a}} \\ \hat{h} \end{bmatrix} \sim \mathcal{GP} \begin{pmatrix} \begin{bmatrix}  \mu_{\bar{a}}(x)\\ \mu_{\hat{h}}(x) \end{bmatrix}, \begin{bmatrix} K_{\bar{a}}(x,x') &  K_{\bar{a}}(x) T(x')^{\transp} \\
   T(x) K_{\bar{a}}(x) & T(x)K_{\bar{a}}(x,x')T(x')^{\transp}\end{bmatrix} \end{pmatrix}. \normalsize
\end{equation}
That is, with \eqref{eq:conditional_mean} one can directly infer $\bar{a}(x)$ from data of the constrained system, as well as condition the constrained acceleration $\hat{h}(x)$ on prior knowledge of $\bar{a}(x)$. % \eg in Example \ref{sec:example_surface} knowledge of $\bar{a}(x)$ at zero velocity is known a priori (gravitational acceleration).

\paragraph{Knowledge transfer between constraint configurations.} In many systems, altering the constraint configuration $\{A(x,\theta_p), b(x,\theta_p)\}$ to a different known configuration \{\colorrulemix{green}{$A'(x,\theta_p'), b'(x,\theta_p')$}\} does not change $\{\bar{a}, M\}$. For example, imagine taking a mass particle (\cref{ex:example_surface}) from one shape of surface to a different one with the same tribological properties. In this case, it is possible to transfer knowledge in form of $\mathcal{D}$ from one system to the different system using the joint distribution
\begin{equation} \label{eq:transfer_learning_distribution}
   \footnotesize \begin{bmatrix} \hat{h} \\ \colorrulemix{green}{\hat{h}'} \end{bmatrix} \sim \mathcal{GP} \begin{pmatrix} \begin{bmatrix}  \mu_{\hat{h}}(x|\theta_p) \\ \colorrulemix{green}{\mu_{\hat{h}'}(x|\theta_p')} \end{bmatrix}, \begin{bmatrix} T(x|\theta_p)K_{\bar{a}}(x,x')T(x'|\theta_p)^{\transp} &  T(x|\theta_p)K_{\bar{a}}(x,x')\colorrulemix{green}{T'(x'|\theta_p')^{\transp}} \\
   \colorrulemix{green}{T'(x|\theta_p')}K_{\bar{a}}(x,x')T(x'|\theta_p)^{\transp} & \colorrulemix{green}{T'(x|\theta_p')}K_{\bar{a}}(x,x')\colorrulemix{green}{T'(x'|\theta_p')^{\transp}}\end{bmatrix} \end{pmatrix}. \normalsize
\end{equation}
%In case of Example \ref{ex:example_unicycle}, it is oftentimes reasonable to assume that the unconstrained dynamics $a(x)$ are fixed over time. However, if a vehicle drives over different surfaces the friction and damping forces, that is, the non-ideal constraint forces $F_z$, change and in return result in different constrained dynamics. 

%===============================================================================
\section{Experimental Results}
\label{sec:evaluations}
In this section, the properties of the proposed GP$^2$ model are analyzed on the benchmark systems detailed in \cref{ex:example_surface} to \ref{ex:example_duffing}. We compare the GP$^2$ to standard (multi-output) GPs.\footnote{The simulation code is available on: \url{https://github.com/AndReGeist/gp_squared}}

%\subsection{Setup}
%\paragraph{\change{Benchmark}} \label{eq:benchmark}
The system parameters are detailed in Sec.\,S6 of the supplementary material.
%The dynamic and constraint functions underlying the benchmark systems are summarized in Appendix \ref{app:benchmark_table} ~\citep[p.~120, 213]{udwadia2007analytical}. 
%For \cref{sec:example_surface}, we set $\theta_p=[0.08, 0.05, 0.05, 0.1, 3]$ and $a_0=0.2$.
%
The input training data consists of randomly sampled observations lying inside the constrained state space. 
%This procedure has been chosen to reduce a potential bias of the learning result on the sampling method used for obtaining data. 
The training data was generated using the analytic ODE \eqref{eq:dynamics_h}.
%For the generation of data, the systems' constraint equations were used to ensure that all states fulfill the constraint manifold. 
The prediction points originate from an equidistant discretization of the constrained state-space. 
The training data is normalized to have zero mean and standard deviation of one. %In case of the GP\textsuperscript{2} model, such a normalization requires to incorporate the normalization matrices inside the model.

%\paragraph{GP models}
As a first baseline for model comparison, the individual $\ddot{q}_i$ are modelled independently as $\ddot{q}_{i} \sim \mathcal{GP}(0, k_{\text{SE}}(x,x'))$, with squared exponential (SE) covariance function $k_{\text{SE}}(x,x')$.
Further, we compare to a standard multi-output GP model, the \citep{gpy2014} implementation of the LMC \citep{alvarez2012kernels} with matrix $B_i=W_iW_i^{\transp}+I_n\kappa$, $W_i \in \mathbb{R}^{n \times r}$, and $\kappa>0$, reading 
$K(x,x')= B_1 k_{\text{SE}}(x,x') + B_2 k_{\text{bias}}(x,x') + B_3 k_{\text{linear}}(x,x')$. Inhere, $k_{\text{linear}}$ and $k_{\text{bias}}$ denote a linear and bias covariance function respectively \citep{williams2006gaussian}. The model  $K(x,x')= B_1 k_{\text{SE}}(x,x')$ is referred to as 
%intrinsic model of coregionalization (
ICM. The hyperparameters are optimized by maximum likelihood estimation via L-BFGS-b \citep{zhu1997algorithm}. For the GP$^2$, we model $\bar{a}\sim \mathcal{GP}(0, k_{\text{SE}}(x,x'))$, see also Sec.\,S2 of the supplementary material, with the same optimization settings as for the other models. For $\mu_{\bar{a}}\neq0$, the prior mean of $\bar{a}$ is set to $\mu_{\bar{a}}=a(x,\theta_p)$ for \cref{ex:example_surface} and \ref{ex:example_unicycle}, while for \cref{ex:example_duffing} $\mu_{\bar{a}}$ models the linear part of the acceleration induced by dampers and springs. Here, the spring and damping parameters are added to $\theta_p$ and estimated alongside the other parameters.

\begin{table}
\centering
\captionof{table}{Comparison of the normalized GPs' predicted mean RMSE, and maximum constraint error for 10 runs.
For the RMSE, the mean, min.\ (subscript) and max.\ (superscript) values are shown.  \label{tab:results_comparison2}}
\footnotesize
\setlength{\tabcolsep}{3.1pt}
\renewcommand{\arraystretch}{1.2}
\begin{tabular*}{0.74\textwidth}{l|lll|lll}
\hline
                    & \multicolumn{3}{l|}{RMSE}  & \multicolumn{3}{l}{max. constraint error} \\ \hline
                    & Surface & Unicycle & Duffing & Surface & Unicycle & Duffing \\ \hline
Analy.\,ODE & --- & --- & --- & 
$2$$\cdot$$10^{-15}$ & $6$$\cdot$$10^{-17}$ & $2$$\cdot$$10^{-14}$\\ \hline
SE & $.235^{\,.272}_{\,.190}$ & $0.27^{\,0.40}_{\,0.20}$ & $.011^{\,.033}_{\,.004}$ & 
$2.6$ & $0.22$ & $6.4$\\
ICM & $.244^{\,.277}_{\,.223}$ & $0.21^{\,0.27}_{\,0.15}$ & $\bm{.003}^{\,\bm{.005}}_{\,.002}$  & 
$1.5$ & $0.22$ & $0.5$\\
LMC & $.194^{\,.233}_{\,.159}$ & $0.21^{\,0.28}_{\,0.15}$ & $\bm{.003}^{\,.006}_{\,\bm{.001}}$  & 
$1.8$ & $0.26$ & $0.023$\\ \hline
GP$^2$,\,$\theta_{p}$$=$$\theta_{p}^{*}$,\,$\mu_{\bar{a}}$$=$$0$ & $.058^{\,.066}_{\,.045}$ & $0.10^{\,0.20}_{\,0.06}$ & $.007^{\,.019}_{\,.001}$  & $4$$\cdot$$10^{-12}$ & $2$$\cdot$$10^{-12}$ & $\bm{1}$$\bm{\cdot}$$\bm{10^{-8}}$ \\
GP$^2$,\,$\theta_{p}$$=$$\theta_{p}^{*}$,\,$\mu_{\bar{a}}$$\neq$$0$ & $\bm{.023}^{\,\bm{.032}}_{\,\bm{.018}}$ & $\bm{0.08}^{\,\bm{0.15}}_{\,\bm{0.05}}$ & $.005^{\,.029}_{\,.002}$  & $\bm{1}$$\bm{\cdot}$$\bm{10^{-13}}$ & $\bm{6}$$\bm{\cdot}$$\bm{10^{-13}}$ & $3$$\cdot$$10^{-8}$ \\ 
\hline
GP$^2$,\,est.\,$\theta_{p}$,\,$\mu_{\bar{a}}$$=$$0$ & $.065^{\,.071}_{\,.056}$ & $0.13^{\,0.30}_{\,\bm{0.05}}$ & $.009^{\,.028}_{\,.003}$  & $0.12$ & $9$$\cdot$$10^{-13}$ & $0.013$ \\
GP$^2$,\,est.\,$\theta_{p}$,\,$\mu_{\bar{a}}$$\neq$$0$ & $.027^{\,.037}_{\,.020}$ & $0.12^{\,0.20}_{\,0.06}$ & $.020^{\,.077}_{\,.004}$  & $0.09$ & $9$$\cdot$$10^{-13}$ & $0.006$ \\ \hline
\end{tabular*}
\normalsize
\vspace{-0.2cm}
\end{table}

\paragraph{Optimization and prediction} 
In each of 10 optimization runs, 100 observations are sampled while the optimization is restarted $30$ times for the benchmark GPs and \change{five} times for the GP$^2$ model. % using random initial values for $\theta$. 
%The differing number of restarts resides in the GPy implementation being much faster than our GP\textsuperscript{2} implementation. 
The prediction results after optimization are depicted in \cref{tab:results_comparison2}. %Inhere we consider the RMSE of the GPs mean predictions at each run and their maximum absolute error of the constraining equation \eqref{eq:constraint}. For \cref{ex:example_surface} and \ref{ex:example_unicycle} if $\mu_{\bar{a}}\neq0$ 
For the mechanically constrained systems of \cref{ex:example_surface} and \ref{ex:example_unicycle}, the GP\textsuperscript{2} shows improved prediction accuracy. The performance can be further increased via the incorporation of additional structural knowledge in form of $\mu_{\bar{a}}$ ($\mu_{\bar{a}} \neq 0$ in \cref{tab:results_comparison2}). If $\theta_p$ is estimated (est.\,$\theta_p$) the constraint error increases. %Albeit, the constraint error is significantly smaller compared to the untransformed models. 
For \cref{ex:example_surface} and \ref{ex:example_duffing}, $\theta_p$ was estimated accurately, whereas the unicycle's parameters ($I_c, R$) converged to their correct ratio.
%For the control-constrained Duffing oscillator, $\theta_p$ converged in some runs to a sub-optimal value, which we reason to the largest function non-linearities (cubic damping) lying in $\bar{a}(x)$ and the limited number of optimization restarts. Yet, for both GP\textsuperscript{2}, the benefit of leveraging {A,B,M} in the model's design is apparent from the results.
In the case of \cref{ex:example_duffing}, a controller constrains the two masses to oscillate synchronously over time. Unlike the unconstrained dynamics that show nonlinear oscillatory behavior, the constrained system dynamics move similar to a single linearly damped oscillator. In this scenario, the GP\textsuperscript{2} compares less favorable to the other models as it is learning on the more complex unconstrained dynamics. For all examples, the GP\textsuperscript{2} demonstrates a considerable improvement with regards to constraint satisfaction.

\begin{figure}[t!]
\centering     %%% not \center
\subfigure[]{\label{fig:a}\includegraphics[width=.32\textwidth]{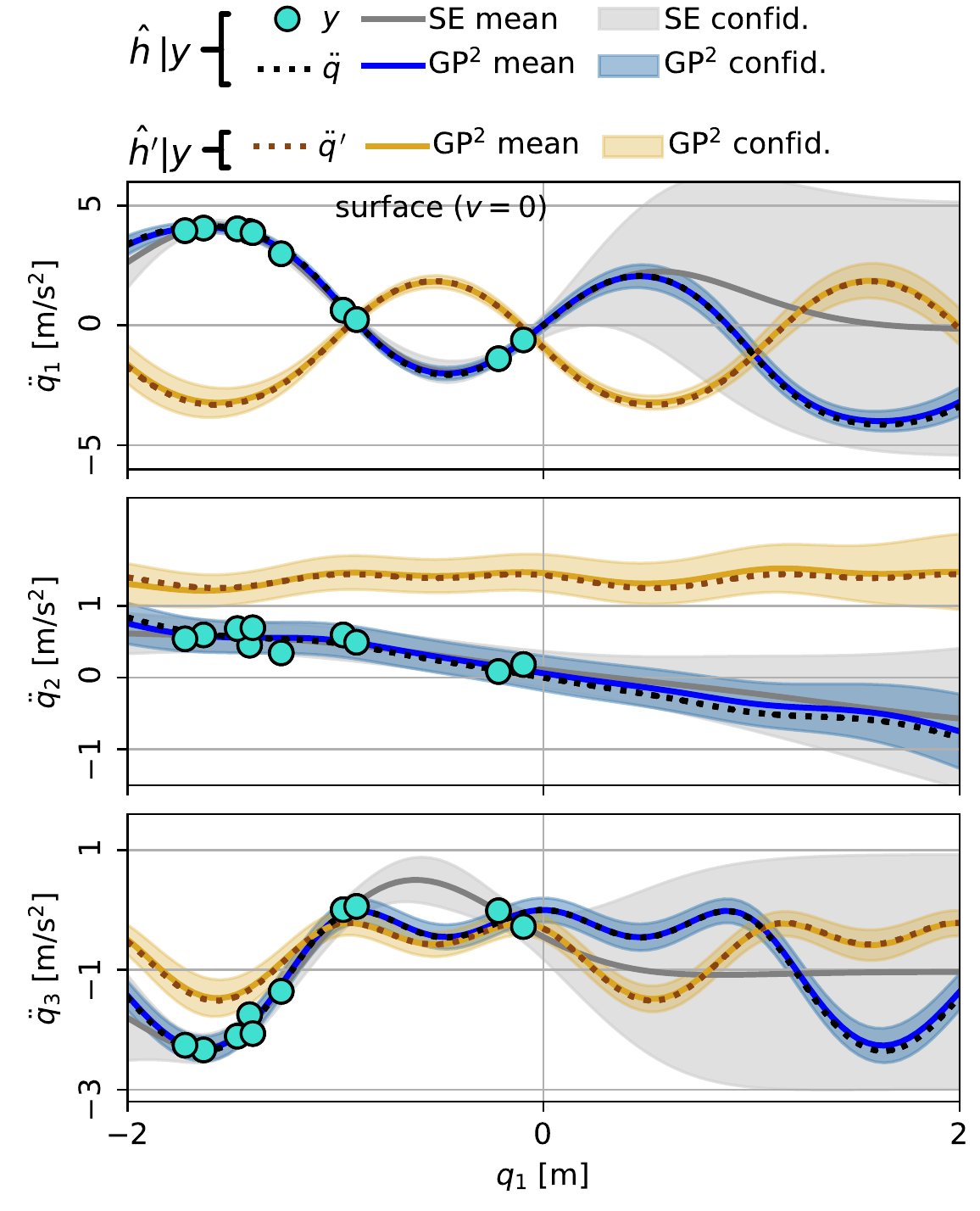}}
\subfigure[]{\label{fig:b}
\begin{minipage}{.32\textwidth}
\centering
\includegraphics[width=1\textwidth]{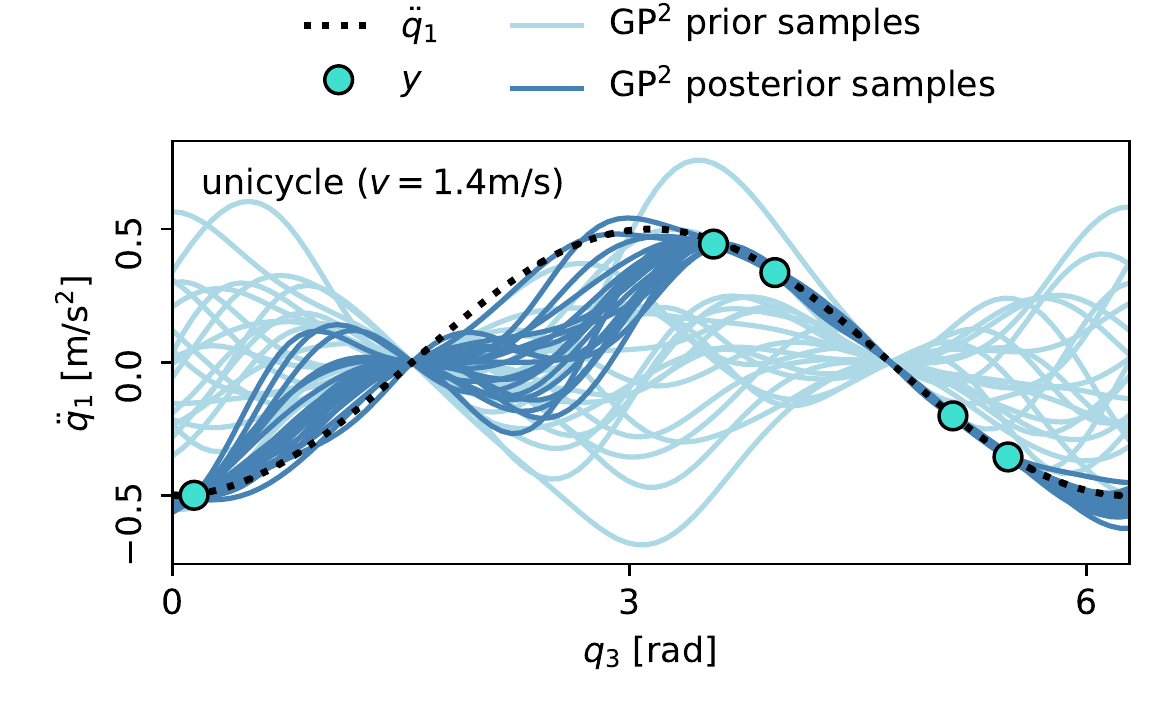}
\includegraphics[width=1\textwidth]{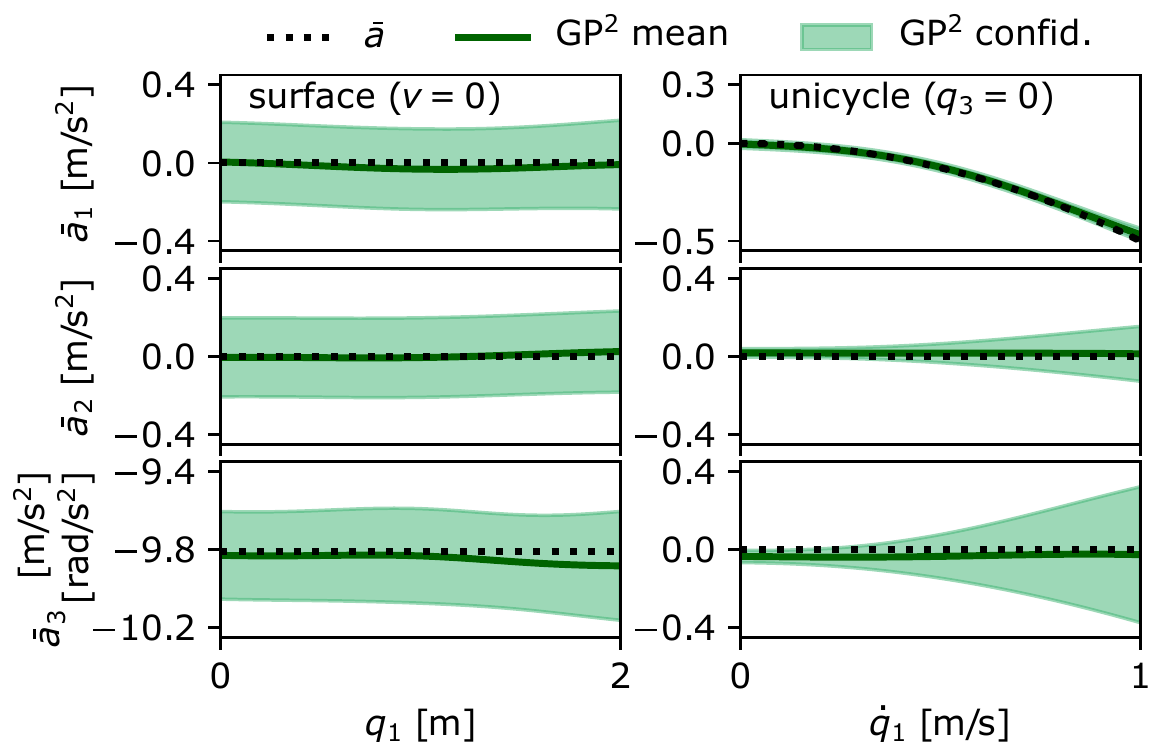}
\end{minipage}
}
\subfigure[]{\label{fig:c}\includegraphics[width=.32\textwidth]{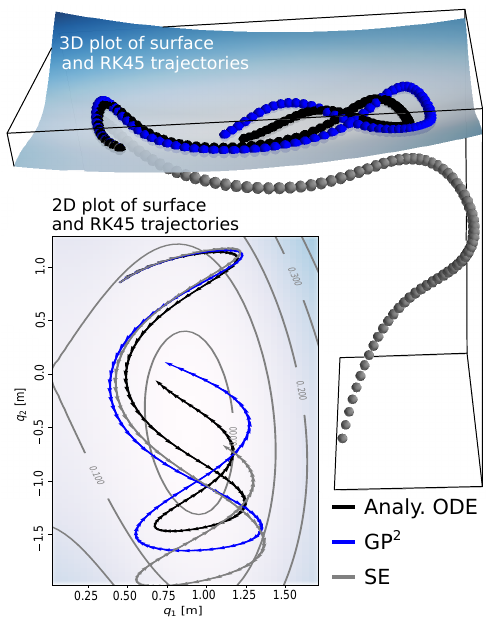}}
\caption{GP\textsuperscript{2} predictions. Figure~\ref{fig:a}: Given $y$ from one surface (blue), predictions of $\ddot{q}$ on the same and $\ddot{q}^{\prime}$ on another surface (yellow). Figure~\ref{fig:b}: (Top) Samples of $\hat{h}$ before and after conditioning on data $y$; (Bottom) Prediction of $\bar{a}|y$. Figure~\ref{fig:c}: RK45 trajectory predictions of \cref{ex:example_surface}.
}
\vspace{-0.2cm}
\end{figure}

%\paragraph{Data Efficiency \& Extrapolation}
%\paragraph{Inferring the Unconstrained Acceleration}
%\paragraph{Transfer Learning}
\paragraph{Extrapolation and transfer}
For illustration of the prediction characteristics, we assume the constraint parameters as given and estimated the remaining GP's hyperparameters on 200 observations. Figure \ref{fig:a} illustrates that the GP\textsuperscript{2} model \emph{extrapolates} the prediction result for \cref{ex:example_surface} ($v=0$). Yet, extrapolation requires $\bar{a}(X)=\bar{a}(x^{*})$. This is not the case for velocity input dimensions when damping plays a predominant role. %, and in case of the Duffing's oscillator only holds with respect to time.
For a different surface $q_3=0.1q_1-0.15q_2-0.1\cos(3q_1)$ resulting in $\hat{h}^{\prime}(x)$ and with $\bar{a}(x)=\bar{a}^{\prime}(x)$, \eqref{eq:transfer_learning_distribution} enables the transfer of the knowledge inherent in $y(x)$ to $\hat{h}^{\prime}(x)$. 
Figure \ref{fig:b} (top) illustrates how the GP\textsuperscript{2}'s samples of $\ddot{q}_{1}(x)$ at $q_3$\,=\,$\ang{90}$ and $q_3$\,=\,$-\ang{90}$ are forced to zero as the unicycle can only translate in driving direction. 
%Inhere, the samples' constraint errors did not exceed $10^{-6}$. 
%We believe that the sample constraint error is larger compared to the mean's constraint error as drawing samples requiries the computation of the Cholesky decomposition of the covariance matrix which required the addition of a small regularization term ($10^{-14}$).
%\paragraph{Prediction of $\bar{a}$} 
Figure \cref{fig:b} (bottom), shows the posterior distribution $\bar{a}|y$ using \eqref{eq:joint distribution2}. For \cref{ex:example_surface} with $v=0$ and $u=0$, $\bar{a}_3$ is simply $g=9.81 \frac{m}{s^2}$. For \cref{ex:example_unicycle} with $u=0$, $\bar{a}$ solely consists of a damping force that increases with the translatory velocity.
%If  $g$ is known, the friction-damping forces can be identified using $y$ and $\{A,b,M\}$ while for the unicycle $\hat{h}(x)$ can be conditioned on prior knowledge, \eg $\bar{a}(v=0)=0$.
%Further, if the constraint forces are solely induced by a controller and an enforced trajectory constraint is determined a posteriori to the experiment, one could potentially infer the unconstrained dynamics from data of the controlled system. As our framework also enables transfer of data between differing constraint configurations one could also use data from different trajectories for such an inference.

\paragraph{Trajectory prediction} Figure \ref{fig:c} illustrates trajectory predictions on \cref{ex:example_surface} computed by a Runke-Kutta-45 (RK45) ODE solver. The solver uses either \eqref{eq:dynamics_h}, the SE model, or the GP\textsuperscript{2} model. 
%The increasing distance of the SE trajectory prediction to the surface leads to a deterioration of the overall prediction performance.
While the SE trajectory prediction leaves the surface, the GP\textsuperscript{2}'s prediction remains on the surface independent of the overall prediction performance. Inhere, the GP\textsuperscript{2} predictions' Euclidean error to the surface increases in the same order of magnitude as with the analytical ODE. In  Sec.\,S7 of the supplementary material, we illustrate that the trajectory predictions of the GP\textsuperscript{2} with estimated hyper-parameters also show improved constraint integrity.

%===============================================================================
\section{Concluding Remarks}
\label{sec:conclusion}
%In this paper, 
We propose a new GP model for learning Lagrangian dynamics that are subject to a non-ideal equality constraint. We leverage that the constraint equation and the system's mass matrix are often straightforward to obtain from a prior mechanical analysis while the constrained dynamics -- with non-ideal forces acting on the system  -- require considerable effort to be modeled parametrically. 
%The proposed method is attractive for model learning of mechanical systems.
For typical mechanical examples, the numerical results demonstrate improved data efficiency and constraint satisfaction. 
%
%While the results and analysis herein show the potential of this approach for combining structural knowledge with non-parametric learning, this work also points to several aspects for further investigations. 
While the numerical results herein treat low-dimensional examples chosen for the purpose of illustration, the method also applies when several constraints act onto the system. Investigating the model's benefits on high-dimensional and hardware experiments is subject of ongoing work, and likewise further analysis of the cases of a singular $M(x)$.  %and $\text{rank}(A(x))<m$. 
%Inhere, we focus on the utility of the GP\textsuperscript{2} model for nonlinear control as well as transfer learning between differing constraint configurations.
%A thorough analysis of the model's theoretical properties if $M(x)$ is singular and $\text{rank}(A(x))<m$ is planned in future.

%In this paper, we propose a new GP model for learning of second-order differential equations with known linear constraints on the second derivative of the integration variable.  As such problems often arise in mechanical systems, where the constraints are obtained from geometric or kinematic considerations, but the exact dynamics are unknown, the proposed method is attractive for model learning of mechanical systems.  Indeed, for typical mechanics examples, the numerical results demonstrate improved data efficiency and extrapolation performance.
%While the results and analysis herein show the potential of this approach for combining structural knowledge with non-parametric learning, this work also points to several aspects for further investigations. While the numerical results herein treat low-dimensional examples chosen for the purpose of illustration, the method equally applies to higher-dimensional problems.  Investigating the methods benefits on problems with more dimensions, 
%as well as in hardware experiments, is subject of ongoing work. Further we will analyze the utility of the GP model for nonlinear control as well as transfer learning between differing constraint configurations.

\acks{The authors thank F.~Solowjow, A.~von Rohr, H.~Haresamudram, and C.~Fiedler for the helpful discussions. We further thank the Cyber Valley Initiative and the international Max Planck research school for intelligent systems (IMPRS-IS) for supporting A.\ Ren\'e Geist.}

%\acks{The authors thank F.~Solowjow, A.~von Rohr, H.~Haresamudram, and C.~Fiedler for the helpful %discussions, as well as the Cyber Valley Initiative and the IMPRS-IS for supporting this work.}

\bibliography{literature}


\section*{Supplementary material}

# Supplementary Material: Learning Constrained Dynamics with Gauss' Principle adhering Gaussian Processes

**A. René Geist** [†]                                      GEIST@IS.MPG.DE

**Sebastian Trimpe** [†]                               TRIMPE@IS.MPG.DE

[†] *Intelligent Control Systems Group, Max Planck*

On the following pages, several supplementary aspects related to the Udwadia-Kalaba equation (UKE) for non-ideally constrained dynamical systems as well as Gauss' Principle adhering Gaussian Processes (GP²) are discussed.

Section S1 details closedness of Gaussian Processes (GPs) under linear operations. Section S2 provides a description on the implementation of the Gauss' principle adhering Gaussian process (GP²) model. Section S3 gives a brief introduction to the role of constraint equations in classical mechanics. Section S4 sketches the derivation of the Udwadia-Kalaba equation for non-ideally constrained systems. Section S5 briefly discusses the case of having a singular mass matrix in the unconstrained equation of motion. Section S6 provides further details on the unicycle system (Example 2) and the controlled Duffing's osscillator system (Example 3). Section S7 further discusses the trajectory prediction results shown in Figure 4.c of the main manuscript.

## S1. Gaussian Process under Linear Operations

While Gaussian distributions are closed under linear transformations, Gaussian processes are closed under linear operations (Rasmussen and Williams, 2006). By applying a functional $\mathcal{A}_x$ on both the mean and covariance function of a GP, its transformed prior distribution is given by

$$\mathcal{A}_x f \sim \mathcal{GP}\big(\mathcal{A}_x \mu(x), \text{Cov}[\mathcal{A}_x f(x), \mathcal{A}_{x'} f(x')]\big), \tag{1}$$

with the transformed covariance matrix reading

$$\begin{aligned}
\text{Cov}[\mathcal{A}_x f(x), \mathcal{A}_{x'} f(x')] &= \mathbb{E}\Big[\big(\mathcal{A}_x f(x) - \mathcal{A}_x \mu(x)\big)\big(\mathcal{A}_{x'} f(x') - \mathcal{A}_{x'} \mu(x')\big)^{\mathsf{T}}\Big] \\
&= \mathcal{A}_x \mathbb{E}\Big[\big(f(x) - \mu(x)\big)\big(f(x') - \mu(x')\big)^{\mathsf{T}}\Big]\mathcal{A}_{x'}^{\mathsf{T}} \\
&= \mathcal{A}_x K(x, x')\mathcal{A}_{x'}^{\mathsf{T}}. \tag{2}
\end{aligned}$$

Note that for the transformed GP to exist, the covariance function $K(x, x')$ must be well defined under the operator $\mathcal{A}_x$. For example if $\mathcal{A}_x$ is a differential operator, then $K(x, x')$ must be a differentiable function. In the Gauss Principle adhering Gaussian Process (GP²) model, the functionals transforming the GP are projection matrices and therefore do not impose additional requirements on the mean and covariance function of the untransformed GP.

## S2. Implementation of GP²

To implement the GP² model, $K_{\bar{a}}(x, x')$ must be chosen. The mean and covariance of the GP² model are then obtained after point-wise transformation of $K_{\bar{a}}(x, x')$ at $\{x_k, x'_k\}$ using $\{A, b, M\}$.

With this, standard GP regression and hyperparameter optimization can be performed. If $A$ is a non-zero row-vector then the computation of the MP-inverse reduces to $A^+ = A^\mathsf{T}/(AA^\mathsf{T})$, whereas if $m > 1$, (Udwadia and Kalaba, 2007, p. 51) suggests a recursive scheme (Greville, 1960) to compute $A^+$. Note that in our implementation, we scaled the outputs of the GP$^2$ model using the empirical mean vector $\mu_s$ and diagonal scaling matrix $S$ containing the standard deviations $\sigma_{s,i}$ of the training observation vectors $y_k, k = 1, ..., N$, such that

$$\hat{h}_s \sim \mathcal{GP}\big(S^{-1}(\mu_{\hat{h}} - \mu_s), S^{-1}k_{\hat{h}}(x, x')(S^{-1})^T\big). \tag{3}$$

## S3. Virtual Displacements and D'Alembert's Principle

Virtual displacements and D'Alembert's Principle are central to understanding the intricacies of constrained motion in mechanical systems. D'Alembert's Principle forms the base for the derivations of all descriptions of constrained dynamical systems in classical mechanics. We mostly omit the arguments of the various functions for the sake of brevity.

Assume that the configuration of a system of rigid bodies is described by the generalized coordinate $n$-vector $q(t)$. The unconstrained motion of the system is expressed by a second-order ODE of the form

$$a(q, \dot{q}, t) = M^{-1}(q, t)^{-1}F_a(q, \dot{q}, t), \tag{4}$$

where $F_a(q, \dot{q}, t)$ denotes an impressed force and $M(q, t)$ a positive-definite and symmetric matrix. Constraints apply an additional force $F_c(q, \dot{q}, t))$ onto the system as a reaction to its current configuration such that the constrained motion of the system is described by

$$\ddot{q} = h(q, \dot{q}, t) = M^{-1}(q, t)\big(F_a(q, \dot{q}, t) + F_c(q, \dot{q}, t)\big). \tag{5}$$

The constraining equation is obtained after differentiating the (non-)holonomic constraint equations (twice) with respect to time such that they take the form

$$A(q, \dot{q}, t)\ddot{q} = b(q, \dot{q}, t). \tag{6}$$

A virtual displacement $\delta q$ denotes the difference between the current displacement at time $t$ and a possible displacement at the *same* time $t$ (Udwadia and Kalaba, 2007, p. 133). The concept of virtual displacements is of enormous importance for classical mechanics as many mechanical constraints impress *solely* constraint forces onto the system such that

$$\delta q^T F_{c,\text{ideal}} = 0. \tag{7}$$

That is the work of these constraint forces is zero under virtual displacements. Equation (7) is referred to as the D'Alembert(-Lagrange)'s principle (d'Alembert, 1743; Lagrange, 1787). Udwadia et al. (1997) extended the discussion on what constitutes a virtual displacement. After analyzing the possible displacement of the constrained system at time $t$ via a Taylor series expansion they concluded that the virtual displacement $\delta q$ fulfills

$$A(q, \dot{q}, t)\delta q = 0. \tag{8}$$

With (7) and the definition of $\delta q$ as in (8), this extended D'Alembert's principle highlights that $F_{c,\text{ideal}} \in \mathcal{R}(A)$, with $\mathcal{R}(A)$ denoting the range space of $A$.

In many mechanical systems the constraint forces $F_c$ *do* work. Therefore, (Udwadia and Kalaba, 2000) proposes a generalization of D'Alembert's principle such that

$$\delta q^T F_c = \delta q^T F_{c,\text{nonideal}}, \tag{9}$$

where $F_{c,\text{nonideal}} \hat{=} F_z$ denotes the part of $F_c = F_{c,\text{ideal}} + F_{c,\text{nonideal}}$ that is coaxial to $\delta q$. With (8), the generalization of D'Alembert's principle states that in addition to $F_{c,\text{ideal}}$, $F_c$ contains a nonideal part $F_z \in \mathcal{N}(A)$, with $\mathcal{N}(A)$ denoting the null space of $A$.

## S4. Udwadia Kalaba Equation with Non-ideal Constraints

The equation of motion for a system being subject to non-ideal constraints is derived in Udwadia and Kalaba (2002). In this derivation the matrix factorization $M^{1/2} = W \Lambda^{1/2} W^T$ with $\Lambda = \text{Diag}(\lambda_1^{1/2}, ... \lambda_n^{1/2})$ and $\lambda_i^{1/2}$ being the $i$-th eigenvalue of $M$ and $W$ containing the eigenvectors of $M$ is used to scale the acceleration of the system such that $\ddot{q}^s = M^{1/2}\ddot{q}$, $a^s = M^{1/2}a$, and $\ddot{q}_c^s = M^{1/2}\ddot{q}_c = M^{-1/2}F_c$. In return the scaled constraining equation (6) reads

$$AM^{-1/2}M^{1/2}\ddot{q} = B\ddot{q}^s = b, \tag{10}$$

and hence the constrained acceleration can be described by the orthogonal decomposition

$$\ddot{q}^s = (B^+B)\ddot{q}^s + (I - B^+B)\ddot{q}^s. \tag{11}$$

Inserting (5) and (10) into (11) yields

$$\ddot{q}^s = B^+b + (I - B^+B)(a^s + \ddot{q}_c^s). \tag{12}$$

From (12) it follows that $\ddot{q}_c^s$ is given by

$$\ddot{q}_c^s = B^+(b - Ba^s) + (I - B^+B)\ddot{q}_c^s, \tag{13}$$

From (9) it follows that the only part of $\ddot{q}_c^s$ that lies in $\mathcal{N}(B)$ is $z^s = M^{-1/2}F_z$, and hence

$$\ddot{q}_c^s = B^+(b - Ba^s) + (I - B^+B)z^s. \tag{14}$$

By use of $B^+ = B^T(BB^T)^+ = (B^TB)^+B^T$ and (14) inserted into (12) the UKE for a nonideally constrained system is given by

$$\begin{aligned}
\ddot{q} &= a + M^{-1/2}B^+(b - Aa) + M^{-1/2}(I - B^+B)z^s, \\
&= M^{-1}A^{\mathsf{T}}(AM^{-1}A^{\mathsf{T}})^+b + (I - M^{-1}A^{\mathsf{T}}(AM^{-1}A^{\mathsf{T}})^+A)(a + z). \tag{15}
\end{aligned}$$

## S5. More on the Rank of the Constraining and Mass Matrices

In this work, we assumed that the mass matrix $M$ of the dynamical system is positive definite.

If the unconstrained acceleration (4) is described by a minimum number of coordinates the Lagrange equations yield a positive definite $M$. While it is in general possible to describe the motion of a mechanical system using a minimum number of coordinates, it can be more practical for the derivation of the unconstrained equation of motion to use more than the minimum number of coordinates. However, this would then result in a singular mass-matrix such that the inverses $M^{-1/2}$ and $M^{-1}$ do not exist and hence (15) is not well defined. Note that Udwadia and Wanichanon (2013) propose an extension of the UKE for nonideal constraint systems with a singular mass matrix.

## S6. Details on the Dynamical Systems of Example 2 and 3

In this section, we further discuss the unicycle and Duffing's oscillator dynamical systems introduced in the experimental results section of the paper. The derivation of these systems is further detailed in (Udwadia and Kalaba, 2007, p. 120, 213).

In general, the functions of the constraining equation $\{A, b\}$ are straightforwardly obtained by a prior kinematic analysis in which first the constraint equations are derived and then differentiated with respect to time. The parameters of the constraining equations can be estimated alongside the GP's parameters. The inertia matrix $M$ is obtained by a rigid body dynamic analysis without the need to consider any induced forces, as these shall be inferred by a GP.

The individual components of the dynamical equations of the unicycle and Duffing's oscillator are depicted in Table 1. The constrained dynamics function of these systems is obtained by inserting the functions in Table 1 together with the non-ideal constraint forces $F_z$ into the UKE.

Fig. 1 to 4 depict the positions and accelerations of the unconstrained as well constrained Duffing's oscillator starting from the initial position $\{q_1(0) = 1, q_2(0) = 1, \dot{q}_1(0) = 2\pi + 2, \dot{q}_2(0) = 2\}$.

Table 1: Mechanical functions of the unconstrained system and the imposed constraints.

| Unicycle | Duffing's Oscillator |
|---|---|
| $M = \begin{bmatrix} m & 0 & -mR\sin(q_3) \\ 0 & m & mR\cos(q_3) \\ -mR\sin(q_3) & mR\cos(q_3) & Ic \end{bmatrix}$ | $M = \begin{bmatrix} m & 0 \\ 0 & m \end{bmatrix}$ |
| $F_a = \begin{bmatrix} mR\dot{q}_3^2\cos(q_3) + \cos(q_3)u_1, \\ mR\dot{q}_3^2\sin(q_3) + \sin(q_3)u_1, \\ u_2 \end{bmatrix}$ | $F_a = K\begin{bmatrix} q_1 \\ q_2 \end{bmatrix} + C\begin{bmatrix} \dot{q}_1 \\ \dot{q}_2 \end{bmatrix} + \begin{bmatrix} k_1^{nl}(q_1 - q_2)^3 \\ k_2^{nl}q_2^3 - k_1^{nl}(q_1 - q_2)^3 \end{bmatrix}$ with $K = \begin{bmatrix} k_1 & -k_1 \\ -k_1 & k_1 + k_2 \end{bmatrix}, C = \begin{bmatrix} c_1 & -c_1 \\ -c_1 & c_1 + c_2 \end{bmatrix}$ |
| $A = \begin{bmatrix} \tan(q_3)\cos(q_3)^2 & -\cos(q_3)^2 & 0 \end{bmatrix}$ | $A = \begin{bmatrix} 1 & -1 \end{bmatrix}$ |
| $b = -\dot{q}_1^2\dot{q}_3$ | $b = -p_1\exp(-p_2q_3)(p_3^2\sin(p_3q_3)$ $+2p_2p_3\cos(p_3t) - p_2^2\sin(p_3q_3)$ |

The parameters chosen for the simulation of the system parameters are denoted in Table 2. The subscripts 'min' and 'max' denote the minimum and maximum state space dimension respectively. Note that in case of the Duffing's oscillator with a mechanistic mean function ($\mu_{\theta_p} \neq 0$) the estimated hyperparameters consist of the constraint parameters, the linear stiffness parameters $k_i^{\text{lin}}$, and linear damping parameters $c_i^{\text{lin}}$.

## S7. Trajectory Generation with GP²

For further illustration of the RK45 prediction in Figure 4 of the main manuscript, Fig. 5 illustrates different trajectory estimation results using differing prediction models. Note that the GP² model with constraint parameter estimation compares favorable to a squared exponential (SE) GP model.

Table 2: Parameters of the system examples in SI-units.

| | Mass on surface | Unicycle | Duffing's oscillator |
|---|---|---|---|
| State space dim. | $\dot{q}_{1,\min} = \dot{q}_{2,\min} = 0,$ <br> $\dot{q}_{1,\max} = \dot{q}_{2,\max} = 1,$ <br> $q_{3,\min} = 0,\ q_{3,\max} = 2\pi,$ <br> $\dot{q}_{3,\min} = -0.5,\ \dot{q}_{3,\max} = 0.5$ <br> $u_{1,\min} = -1,\ u_{1,\max} = 1,$ <br> $u_{2,\min} = -0.5,\ u_{2,\max} = 0.5$ | $\dot{q}_{1,\min} = \dot{q}_{2,\min} = 0,$ <br> $\dot{q}_{1,\max} = \dot{q}_{2,\max} = 1,$ <br> $q_{3,\min} = 0,\ q_{3,\max} = 2\pi,$ <br> $\dot{q}_{3,\min} = -0.5,\ \dot{q}_{3,\max} = 0.5$ <br> $u_{1,\min} = -1,\ u_{1,\max} = 1,$ <br> $u_{2,\min} = -0.5,\ u_{2,\max} = 0.5$ | $q_{1,\min} = q_{2,\min} = -4,$ <br> $q_{1,\max} = q_{2,\max} = 4,$ <br> $\dot{q}_{1,\min} = \dot{q}_{2,\min} = -5,$ <br> $\dot{q}_{1,\max} = \dot{q}_{2,\max} = 5,$ <br> $t_{\max} = 0,\ t_{\max} = 5,$ |
| System param. | $m = 3,\ g = -9.81,\ a_0 = 0.2,$ <br> $\theta_p = [0.08, 0.05, 0.05, 0.1, 3]$ | $m = 1,\ a_0 = 0.5,$ <br> $R = 0.05,\ I_c = 0.02,$ <br> $\theta_p = [R, I_c]$ | $m_1 = 2,\ m_2 = 1,$ <br> $k_1^{\mathrm{nl}} = 1,\ k_2^{\mathrm{nl}} = 2,$ <br> $k_1^{\mathrm{lin}} = 10,\ k_2^{\mathrm{lin}} = 12,$ <br> $c_1^{\mathrm{lin}} = 0.1,\ c_2^{\mathrm{lin}} = 0.15,$ <br> $p_1 = 1,\ p_2 = 0.3,\ p_3 = 2\pi$ |

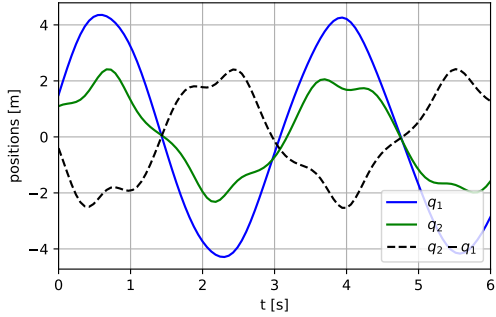

Figure 1: Unconstrained Duffing's oscillator's position plotted over time.

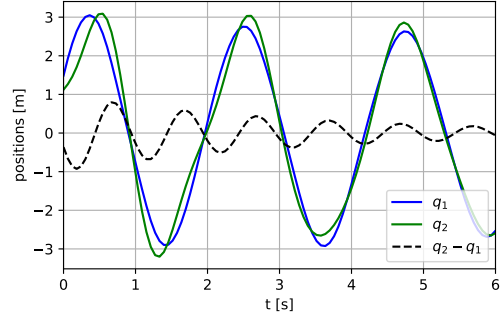

Figure 2: Constrained Duffing's oscillator's position plotted over time.

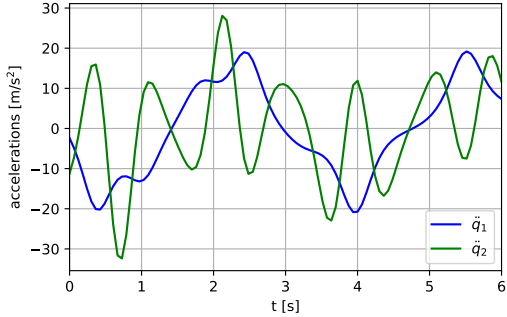

Figure 3: Unconstrained Duffing's oscillator's accelerations plotted over time.

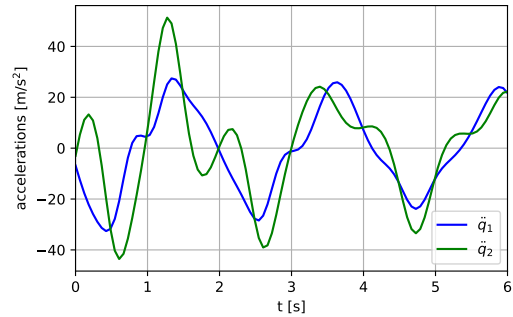

Figure 4: Constrained Duffing's oscillator's accelerations plotted over time.


\begin{minipage}[t]{1\linewidth}
\includepdf[pages=6]{s5_ArxivVersion_L4DC_supplementary.pdf}
\end{minipage}
\end{document}

%% file: fig/tangent_on_surface_small2.pdf_tex
%% Accompanies image file 'tangent_on_surface_small2.pdf' (pdf, eps, ps)
\begingroup%
  \makeatletter%
  \providecommand\color[2][]{%
    \errmessage{(Inkscape) Color is used for the text in Inkscape, but the package 'color.sty' is not loaded}%
    \renewcommand\color[2][]{}%
  }%
  \providecommand\transparent[1]{%
    \errmessage{(Inkscape) Transparency is used (non-zero) for the text in Inkscape, but the package 'transparent.sty' is not loaded}%
    \renewcommand\transparent[1]{}%
  }%
  \providecommand\rotatebox[2]{#2}%
  \newcommand*\fsize{\dimexpr\f@size pt\relax}%
  \newcommand*\lineheight[1]{\fontsize{\fsize}{#1\fsize}\selectfont}%
  \ifx\svgwidth\undefined%
    \setlength{\unitlength}{164.35284111bp}%
    \ifx\svgscale\undefined%
      \relax%
    \else%
      \setlength{\unitlength}{\unitlength * \real{\svgscale}}%
    \fi%
  \else%
    \setlength{\unitlength}{\svgwidth}%
  \fi%
  \global\let\svgwidth\undefined%
  \global\let\svgscale\undefined%
  \makeatother%
  \begin{picture}(1,0.90196792)%
    \lineheight{1}%
    \setlength\tabcolsep{0pt}%
    \put(0,0){\includegraphics[width=\unitlength,page=1]{tangent_on_surface_small2.pdf}}%
    \put(0.0189028,0.15978535){\color[rgb]{0,0,0}\makebox(0,0)[lt]{\begin{minipage}{0.8521363\unitlength}\raggedright surface\end{minipage}}}%
    \put(0,0){\includegraphics[width=\unitlength,page=2]{tangent_on_surface_small2.pdf}}%
    \put(0.59114673,0.72279148){\color[rgb]{0,0,0}\makebox(0,0)[lt]{\begin{minipage}{1.15201376\unitlength}\raggedright $\text{tangent plane, }\mathcal{N}(A)$\end{minipage}}}%
    \put(0,0){\includegraphics[width=\unitlength,page=3]{tangent_on_surface_small2.pdf}}%
    \put(0.06582495,0.07366676){\color[rgb]{0,0,0}\makebox(0,0)[lt]{\begin{minipage}{0.8521363\unitlength}\raggedright mass particle\end{minipage}}}%
    \put(0,0){\includegraphics[width=\unitlength,page=4]{tangent_on_surface_small2.pdf}}%
    \put(0.48617536,0.81961091){\color[rgb]{0,0,0}\makebox(0,0)[lt]{\begin{minipage}{0.8521363\unitlength}\raggedright \textbf{$a$}\end{minipage}}}%
    \put(0.37026314,0.4170626){\color[rgb]{0,0,0}\makebox(0,0)[lt]{\begin{minipage}{0.8521363\unitlength}\raggedright \textbf{$\ddot{q}$}\end{minipage}}}%
    \put(0.3883991,0.6195157){\color[rgb]{0,0,0}\makebox(0,0)[lt]{\begin{minipage}{0.8521363\unitlength}\raggedright \textbf{$z$}\end{minipage}}}%
    \put(0.22894041,0.808174){\color[rgb]{0,0,0}\makebox(0,0)[lt]{\begin{minipage}{0.8521363\unitlength}\raggedright $-LAa$\end{minipage}}}%
    \put(0,0){\includegraphics[width=\unitlength,page=5]{tangent_on_surface_small2.pdf}}%
    \put(-0.05449018,0.98013042){\color[rgb]{0,0,0}\makebox(0,0)[lt]{\begin{minipage}{1.05726713\unitlength}\raggedright $\tau_{\mathrm{ideal}}$$=$$L(b-Aa)$\end{minipage}}}%
    \put(0.19414247,0.56978695){\color[rgb]{0,0,0}\makebox(0,0)[lt]{\begin{minipage}{0.8521363\unitlength}\raggedright $Lb$\end{minipage}}}%
    \put(0,0){\includegraphics[width=\unitlength,page=6]{tangent_on_surface_small2.pdf}}%
  \end{picture}%
\endgroup%

%% file: fig/unicycle5.pdf_tex
%% Accompanies image file 'unicycle5.pdf' (pdf, eps, ps)
\begingroup%
  \makeatletter%
  \providecommand\color[2][]{%
    \errmessage{(Inkscape) Color is used for the text in Inkscape, but the package 'color.sty' is not loaded}%
    \renewcommand\color[2][]{}%
  }%
  \providecommand\transparent[1]{%
    \errmessage{(Inkscape) Transparency is used (non-zero) for the text in Inkscape, but the package 'transparent.sty' is not loaded}%
    \renewcommand\transparent[1]{}%
  }%
  \providecommand\rotatebox[2]{#2}%
  \newcommand*\fsize{\dimexpr\f@size pt\relax}%
  \newcommand*\lineheight[1]{\fontsize{\fsize}{#1\fsize}\selectfont}%
  \ifx\svgwidth\undefined%
    \setlength{\unitlength}{90.70866142bp}%
    \ifx\svgscale\undefined%
      \relax%
    \else%
      \setlength{\unitlength}{\unitlength * \real{\svgscale}}%
    \fi%
  \else%
    \setlength{\unitlength}{\svgwidth}%
  \fi%
  \global\let\svgwidth\undefined%
  \global\let\svgscale\undefined%
  \makeatother%
  \begin{picture}(1,1.5)%
    \lineheight{1}%
    \setlength\tabcolsep{0pt}%
    \put(0,0){\includegraphics[width=\unitlength,page=1]{unicycle5.pdf}}%
    \put(-0.01311809,0.65436426){\color[rgb]{0,0,0}\makebox(0,0)[lt]{\lineheight{1.25}\smash{\begin{tabular}[t]{l}$q_2$\end{tabular}}}}%
    \put(0.62787014,0.09249464){\color[rgb]{0,0,0}\makebox(0,0)[lt]{\lineheight{1.25}\smash{\begin{tabular}[t]{l}$q_1$\end{tabular}}}}%
    \put(0.00217463,0.00952116){\color[rgb]{0,0,0}\makebox(0,0)[lt]{\lineheight{1.25}\smash{\begin{tabular}[t]{l}$q_3$\end{tabular}}}}%
    \put(0.36346406,0.48755994){\color[rgb]{0,0,0}\makebox(0,0)[lt]{\lineheight{1.25}\smash{\begin{tabular}[t]{l}$C$\end{tabular}}}}%
    \put(0.53230506,0.50841871){\color[rgb]{0,0,0}\makebox(0,0)[lt]{\lineheight{1.25}\smash{\begin{tabular}[t]{l}$G$\end{tabular}}}}%
    \put(0.19699258,0.66300264){\color[rgb]{0,0,0}\makebox(0,0)[lt]{\lineheight{1.25}\smash{\begin{tabular}[t]{l}$R$\end{tabular}}}}%
    \put(0.61443697,1.57533915){\color[rgb]{0,0,0}\makebox(0,0)[lt]{\begin{minipage}{1.93965912\unitlength}\raggedright $\tau_{\mathrm{ideal}}$\end{minipage}}}%
    \put(0.89533019,1.05187852){\color[rgb]{0,0,0}\makebox(0,0)[lt]{\begin{minipage}{1.5439653\unitlength}\raggedright \textbf{$z$}\end{minipage}}}%
    \put(0.59212928,0.88999102){\color[rgb]{0,0,0}\makebox(0,0)[lt]{\begin{minipage}{1.5439653\unitlength}\raggedright \textbf{$\ddot{q}$}\end{minipage}}}%
    \put(0.33478156,1.28289852){\color[rgb]{0,0,0}\makebox(0,0)[lt]{\begin{minipage}{1.5439653\unitlength}\raggedright \textbf{$a$}\end{minipage}}}%
    \put(0,0){\includegraphics[width=\unitlength,page=2]{unicycle5.pdf}}%
    \put(0.82575928,1.32093602){\color[rgb]{0,0,0}\makebox(0,0)[lt]{\begin{minipage}{1.5439653\unitlength}\raggedright $-LAa$\end{minipage}}}%
    \put(0.78979925,0.87289852){\color[rgb]{0,0,0}\makebox(0,0)[lt]{\begin{minipage}{1.5439653\unitlength}\raggedright $Lb$\end{minipage}}}%
    \put(-0.23828197,0.39971227){\color[rgb]{0,0,0}\makebox(0,0)[lt]{\begin{minipage}{1.5439653\unitlength}\raggedright $\mathcal{N}(A)$\end{minipage}}}%
    \put(0,0){\includegraphics[width=\unitlength,page=3]{unicycle5.pdf}}%
  \end{picture}%
\endgroup%

%% file: fig/duffing.pdf_tex
%% Accompanies image file 'duffing.pdf' (pdf, eps, ps)
\begingroup%
  \makeatletter%
  \providecommand\color[2][]{%
    \errmessage{(Inkscape) Color is used for the text in Inkscape, but the package 'color.sty' is not loaded}%
    \renewcommand\color[2][]{}%
  }%
  \providecommand\transparent[1]{%
    \errmessage{(Inkscape) Transparency is used (non-zero) for the text in Inkscape, but the package 'transparent.sty' is not loaded}%
    \renewcommand\transparent[1]{}%
  }%
  \providecommand\rotatebox[2]{#2}%
  \newcommand*\fsize{\dimexpr\f@size pt\relax}%
  \newcommand*\lineheight[1]{\fontsize{\fsize}{#1\fsize}\selectfont}%
  \ifx\svgwidth\undefined%
    \setlength{\unitlength}{170.07874016bp}%
    \ifx\svgscale\undefined%
      \relax%
    \else%
      \setlength{\unitlength}{\unitlength * \real{\svgscale}}%
    \fi%
  \else%
    \setlength{\unitlength}{\svgwidth}%
  \fi%
  \global\let\svgwidth\undefined%
  \global\let\svgscale\undefined%
  \makeatother%
  \begin{picture}(1,0.41666667)%
    \lineheight{1}%
    \setlength\tabcolsep{0pt}%
    \put(0,0){\includegraphics[width=\unitlength,page=1]{duffing.pdf}}%
    \put(0.41584266,0.38145582){\color[rgb]{0,0,0}\makebox(0,0)[lt]{\lineheight{1.25}\smash{\begin{tabular}[t]{l}$q_2$\end{tabular}}}}%
    \put(0.78150425,0.38144666){\color[rgb]{0,0,0}\makebox(0,0)[lt]{\lineheight{1.25}\smash{\begin{tabular}[t]{l}$q_1$\end{tabular}}}}%
    \put(0.32764009,0.14403516){\color[rgb]{0,0,0}\makebox(0,0)[lt]{\lineheight{1.25}\smash{\begin{tabular}[t]{l}$m_2$\end{tabular}}}}%
    \put(0.68272463,0.14414554){\color[rgb]{0,0,0}\makebox(0,0)[lt]{\lineheight{1.25}\smash{\begin{tabular}[t]{l}$m_1$\end{tabular}}}}%
    \put(0.22112182,0.27379692){\color[rgb]{0,0,0}\makebox(0,0)[lt]{\lineheight{1.25}\smash{\begin{tabular}[t]{l}$c_2$\end{tabular}}}}%
    \put(0.1358311,0.02181328){\color[rgb]{0,0,0}\makebox(0,0)[lt]{\lineheight{1.25}\smash{\begin{tabular}[t]{l}$s_2$\end{tabular}}}}%
    \put(0.56892885,0.2848214){\color[rgb]{0,0,0}\makebox(0,0)[lt]{\lineheight{1.25}\smash{\begin{tabular}[t]{l}$c_1$\end{tabular}}}}%
    \put(0.47443479,0.02181277){\color[rgb]{0,0,0}\makebox(0,0)[lt]{\lineheight{1.25}\smash{\begin{tabular}[t]{l}$s_1$\end{tabular}}}}%
  \end{picture}%
\endgroup%